\documentclass[letterpaper]{article} 
\usepackage{aaai2027}  
\usepackage[hyphens]{url}  
\usepackage{graphicx} 
\usepackage{natbib}  
\usepackage{caption} 
\usepackage{algorithm}
\usepackage{algorithmic}
\usepackage{booktabs}
\usepackage{multirow}
\usepackage{graphicx}
\usepackage{amssymb}
\usepackage{graphicx}   
\usepackage{booktabs}   
\usepackage{multirow}   
\usepackage[table]{xcolor} 
\usepackage{amsmath}
\usepackage{graphicx}
\usepackage{booktabs}
\usepackage{multirow}
\usepackage[table]{xcolor}
\definecolor{bestpurple}{RGB}{238,210,255}
\definecolor{secondyellow}{RGB}{255,242,204}
\providecommand{\NA}{N/A}

\newcommand{\overallbest}[1]{\cellcolor{bestpurple}#1}
\newcommand{\overallsecond}[1]{\cellcolor{secondyellow}#1}
\newcommand{\catbest}[1]{\textbf{#1}}
\newcommand{\catsecond}[1]{\underline{#1}}

\providecommand{\NA}{N/A}
\providecommand{\best}[1]{\cellcolor{bestpurple}\textbf{#1}}
\providecommand{\second}[1]{\cellcolor{secondyellow}\underline{#1}}

\usepackage{newfloat}
\usepackage{listings}
\DeclareCaptionStyle{ruled}{labelfont=normalfont,labelsep=colon,strut=off} 
\floatstyle{ruled}
\newfloat{listing}{tb}{lst}{}
\floatname{listing}{Listing}

\usepackage{booktabs}

\title{RbFT-Net: Rectify-Before-Fuse Temporal Radar Anchors for 4D Radar--Camera Depth Completion}
\author{
    Wentao Zhao\textsuperscript{\rm 1},
    Shouxuan Wu\textsuperscript{\rm 2},
    Yongtao Cen\textsuperscript{\rm 1},
    Tianchen Deng\textsuperscript{\rm 1},
    Yuyang Zhang\textsuperscript{\rm 3},
    Jingchuan Wang\textsuperscript{\rm 1}
}

\affiliations{
    \textsuperscript{\rm 1}School of Automation and Intelligent Sensing,
    Institute of Medical Robotics, Shanghai Jiao Tong University,
    Shanghai 200240, China\\
    \textsuperscript{\rm 2}School of Electronic and Information Engineering,
    Beijing Jiaotong University, Beijing, China\\
    \textsuperscript{\rm 3}State Key Laboratory of Advanced Rail Autonomous Operation
    and School of Electronic and Information Engineering,
    Beijing Jiaotong University, Beijing 100044, China
}

\begin{document}

\maketitle

\begin{abstract}
Dense metric depth prediction from cameras and millimeter-wave radar offers a cost-effective sensing solution for autonomous systems. However, radar measurements are inherently sparse and susceptible to clutter, multipath reflections, and projection errors. While aggregating multiple radar frames provides denser metric cues, it also introduces temporal misalignment and dynamic-object interference. Directly propagating such unreliable measurements can therefore corrupt large regions of the predicted depth map. To address this issue, we propose RbFT-Net, an end-to-end \emph{rectify-before-fuse} framework for multi-frame 4D radar--camera depth completion. Rather than assuming accumulated radar returns to be accurate, RbFT-Net treats them as noisy temporal anchor candidates. An image-conditioned rectification module jointly corrects their image-plane locations and metric depths while estimating pointwise reliability. The rectified anchors are then selectively propagated before high-level multi-modal fusion, suppressing the influence of unreliable measurements. Experiments on ZJU-4DRadarCam and a newly collected 4D radar--camera--LiDAR dataset show that RbFT-Net consistently outperforms the evaluated independent radar--camera methods and remains competitive with plug-in pipelines using auxiliary monocular depth models. Cross-platform evaluation and component analyses further support the effectiveness of the proposed rectification and reliability-aware propagation strategy.
\end{abstract}


\section{Introduction}

Dense metric depth estimation is fundamental to autonomous driving,
robotic navigation, and 3D scene understanding.
Cameras provide rich appearance cues but cannot directly resolve metric
scale and may degrade under adverse visual conditions, whereas LiDAR
provides accurate geometry at a relatively high cost.
Millimeter-wave radar offers direct metric ranging and robustness to
illumination changes, making it an attractive complementary sensor for
dense depth prediction
~\cite{lin2020depth,lo2021depth,long2021radar,singh2023depth}.
In particular, 4D radar resolves elevation in addition to range,
azimuth, and Doppler, allowing height-aware radar returns to be
projected into the camera view as sparse metric cues
~\cite{li2024radarcam,sun2024cafnet,wang2025tacodepth}.

Existing radar--camera depth completion methods commonly propagate
projected radar measurements to dense image regions under visual
guidance
~\cite{long2021radar,singh2023depth,li2024radarcam,
sun2024cafnet,sun2025get}.
However, radar returns are extremely sparse and may be inconsistent
with visual structures because of limited angular resolution,
multipath effects, clutter, calibration uncertainty, and radar-to-image
misprojection.
Once propagated, inaccurate radar measurements can contaminate much
larger image regions, particularly around object boundaries, thin
structures, and distant objects.
Recent plug-in methods alleviate the limited structural information
of radar by incorporating dense predictions from separately trained
monocular depth models
~\cite{li2024radarcam,qin2025racalnet,wang2025tacodepth}.
Although these predictions provide strong structural priors, they
introduce an additional model dependency and increase the size and
complexity of the complete inference pipeline.
This motivates an independent solution that directly predicts dense
metric depth from radar and RGB inputs.

A complementary source of information is the temporal measurements
naturally available in radar streams.
Consecutive radar frames provide complementary metric observations
and can alleviate the sparsity of single-frame input.
A conventional temporal fusion strategy is to use ego-motion to warp
historical measurements into the current view \cite{zhou2025manydepth2}.
Although such geometric alignment reduces rigid inter-frame
displacement, it does not guarantee accurate radar-to-image anchors:
the warped returns may still be affected by calibration uncertainty,
radar depth noise, multipath reflections, and independently moving
objects.
Direct accumulation further retains temporal misalignment while also
aggregating clutter and other unreliable returns.
Multi-frame radar can therefore become denser without becoming more
reliable, potentially amplifying errors during subsequent propagation.
This raises the central question of this work:
\emph{how can noisy temporal radar measurements be rectified and
selectively exploited before their errors propagate into dense depth
predictions?}

To address this question, we propose \textbf{RbFT-Net}, an end-to-end
\emph{rectify-before-fuse} framework for multi-frame 4D radar--camera
depth completion.
Rather than assuming accumulated radar returns to be accurate depth
measurements, RbFT-Net treats them as noisy temporal anchor candidates.
An image-conditioned rectification module jointly corrects their
image-plane locations and metric depths while estimating pointwise
reliability.
The resulting anchors are then selectively propagated according to
their reliability and target--anchor compatibility before high-level
multi-modal fusion and dense refinement.
This design suppresses unreliable temporal measurements before their
errors spread over the image.
Operating directly on accumulated radar returns, RbFT-Net predicts
dense metric depth from radar and RGB without relying on an auxiliary
monocular depth model.
Combined with an efficient dense prediction backbone, it forms a
compact independent pipeline.

We conduct extensive experiments on ZJU-4DRadarCam
~\cite{li2024radarcam} and a newly collected dataset acquired using a
different 4D radar--camera platform.
RbFT-Net achieves the strongest overall performance among independent
methods and remains competitive with plug-in pipelines across both
datasets.
Zero-shot and limited-data adaptation experiments further evaluate
its cross-sensor transferability, while comprehensive ablations and
efficiency analyses validate the proposed components and demonstrate
a favorable accuracy--efficiency trade-off.

Our main contributions are summarized as follows:
\begin{itemize}
    \item We propose RbFT-Net, an end-to-end rectify-before-fuse
    framework that rectifies noisy temporal radar anchors before
    reliability-aware propagation and high-level multi-modal fusion.

    \item We introduce an image-conditioned temporal anchor
    rectification module that jointly corrects the image-plane
    locations and metric depths of noisy radar returns while
    estimating their pointwise reliability.

    \item We develop a reliability-aware propagation strategy that
    combines anchor reliability with target--anchor compatibility to
    selectively transfer trustworthy metric evidence to dense image
    regions.

    \item We collect a new 4D radar--camera--LiDAR dataset using a
    different sensing platform and conduct in-domain, zero-shot, and
    limited-data cross-sensor evaluations.
    The dataset and evaluation protocol will be made publicly available.
\end{itemize}

\begin{figure*}[t]
    \centering
    \includegraphics[width=\textwidth]{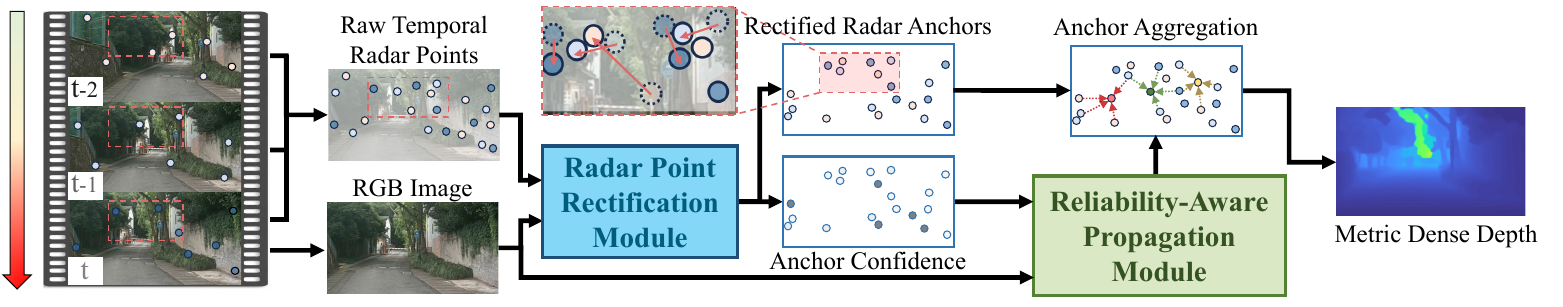}
    \caption{
    Overview of RbFT-Net.
    Directly accumulated multi-frame 4D radar returns are treated as noisy
    temporal anchor candidates.
    Image-conditioned rectification corrects their image-plane locations and
    metric depths while estimating pointwise reliability.
    The resulting anchors are selectively propagated according to their
    reliability and target--anchor compatibility to produce dense metric depth.
    }
    \label{fig:framework}
\end{figure*}

\section{Related Work}

Following TacoDepth~\cite{wang2025tacodepth}, we categorize
radar--camera depth completion methods as independent or plug-in.
Independent methods directly predict metric depth from radar and RGB,
whereas plug-in methods additionally rely on an auxiliary monocular
depth model.
We also review temporal fusion for depth estimation.

\paragraph{Independent Radar--Camera Depth Completion.}

Depth completion commonly propagates sparse measurements using learned
spatial affinities or image-conditioned weights.
Representative methods such as CSPN++~\cite{cheng2020cspn++} and
BP-Net~\cite{tang2024bilateral} are effective for relatively accurate
LiDAR samples, but can be sensitive to the substantially sparser and
noisier measurements provided by radar.

Independent radar--camera methods directly fuse radar and RGB inputs.
Early studies explored sensor complementarity, pixel--depth association,
and ordinal depth prediction
~\cite{lin2020depth,long2021radar,lo2021depth,singh2023depth}.
Subsequent methods improve radar utilization through confidence
modeling~\cite{sun2024cafnet}, sparse supervision~\cite{li2024sparse},
geometric upsampling~\cite{sun2025get}, lightweight
distillation~\cite{sun2025lircdepth,sun2025xd},
transformer-based fusion~\cite{huang2024rcdformer}, and
structure-aware modeling~\cite{zhang2025structure}.
JustDepth~\cite{yun2026justdepth} focuses on efficient estimation under
sparse supervision, while TacoDepth~\cite{wang2025tacodepth} provides
an independent one-stage configuration.
Despite recent efforts to recalibrate projected radar
returns~\cite{qin2025racalnet}, existing methods generally do not jointly
rectify their image locations and metric depths while estimating reliability
before dense propagation.

\paragraph{Plug-in Radar--Camera Depth Completion.}

Plug-in methods combine radar measurements with predictions from a
separately trained monocular depth model.
RadarCam-Depth~\cite{li2024radarcam} incorporates 4D radar into an
auxiliary monocular depth representation, while radar-guided polynomial
fitting~\cite{rim2025radar} and RaCalNet~\cite{qin2025racalnet}
recover the metric scale and shift of affine-invariant depth
predictions.
TacoDepth~\cite{wang2025tacodepth} also supports a plug-in
configuration.
Although these methods benefit from dense monocular structure, their
complete inference pipelines require an additional depth network,
motivating more compact independent alternatives.

\paragraph{Temporal Fusion for Depth Estimation.}

Multi-frame monocular methods exploit neighboring images through
geometric matching, cost volumes, or motion-aware modeling
~\cite{watson2021temporal,guizilini2022multiframe,
zhou2025manydepth2}, typically relying on relative poses or geometric
warping.
Radar--camera methods commonly accumulate consecutive radar sweeps to
increase input density
~\cite{lin2020depth,long2021radar,gasperini2021r4dyn,
lo2021depth}.
However, direct accumulation can introduce temporal misalignment,
dynamic returns, clutter, and multipath interference, making denser
radar input not necessarily more reliable.
In contrast, RbFT-Net treats accumulated returns as noisy temporal
anchors and jointly rectifies their locations and depths while estimating
their reliability before dense propagation.

\section{Method}
\subsection{Overview}
\label{sec:method_overview}

Given the current RGB image $I_t$ and a short temporal window of
4D radar frames $\{\mathcal{R}_{t-k}\}_{k=0}^{T-1}$, our goal is
to predict a dense metric depth map $D_t$ in the current camera
view.
As illustrated in Fig.~\ref{fig:framework}, RbFT-Net follows a
\emph{rectify-before-fuse} paradigm: it first rectifies temporal
radar anchors using image context and then performs
reliability-aware propagation and high-level multi-modal fusion.

Radar returns from all $T$ frames are directly projected into the
current image using the calibrated radar--camera extrinsics and camera
intrinsics, without ego-motion compensation:
\begin{equation}
\mathcal{A}_t
=
\left\{
a_i=
\left(
\mathbf{p}_i,z_i,\mathbf{r}_i
\right)
\right\}_{i=1}^{N},
\end{equation}
where $\mathbf{p}_i=(u_i,v_i)$ is the projected location in the current
image, $z_i$ is the corresponding metric depth, and $\mathbf{r}_i$
contains the radar attributes.
The candidates are processed as an unordered set without explicit
temporal-index encoding.
Although direct accumulation increases observation density, it also
introduces temporal misalignment under sensor or object motion and
retains cluttered returns.
Therefore, $\mathcal{A}_t$ is treated as a set of noisy anchor candidates
rather than reliable depth measurements.

The image-conditioned anchor rectification module samples local
visual context around each candidate and predicts its spatial
offset, depth residual, and pointwise reliability.
The resulting rectified anchor is represented as
\begin{equation}
\hat{a}_i
=
\left(
\hat{\mathbf{p}}_i,
\hat{z}_i,
c_i,
\mathbf{f}_i^a
\right),
\end{equation}
where $\hat{\mathbf{p}}_i=(\hat{u}_i,\hat{v}_i)$ and
$\hat{z}_i$ are the rectified image location and metric depth,
$c_i\in[0,1]$ is the estimated reliability, and
$\mathbf{f}_i^a$ is the anchor feature used for propagation.

The reliability-aware propagation module evaluates nearby
anchors according to their reliability and target--anchor
compatibility, aggregates informative metric evidence, and
performs image-guided refinement to produce the final depth map.
The overall framework is summarized as
\begin{equation}
\hat{\mathcal{A}}_t
=
\mathcal{F}_{\mathrm{rect}}
\left(
\mathcal{A}_t,I_t
\right),
\qquad
D_t
=
\mathcal{F}_{\mathrm{prop}}
\left(
I_t,\mathcal{R}_t,\hat{\mathcal{A}}_t
\right),
\end{equation}
where $\mathcal{F}_{\mathrm{rect}}$ denotes image-conditioned
anchor rectification, while $\mathcal{F}_{\mathrm{prop}}$ includes
reliability-aware propagation, high-level multi-modal fusion, and
dense refinement.

\begin{figure*}[t]
    \centering
    \includegraphics[width=\textwidth]{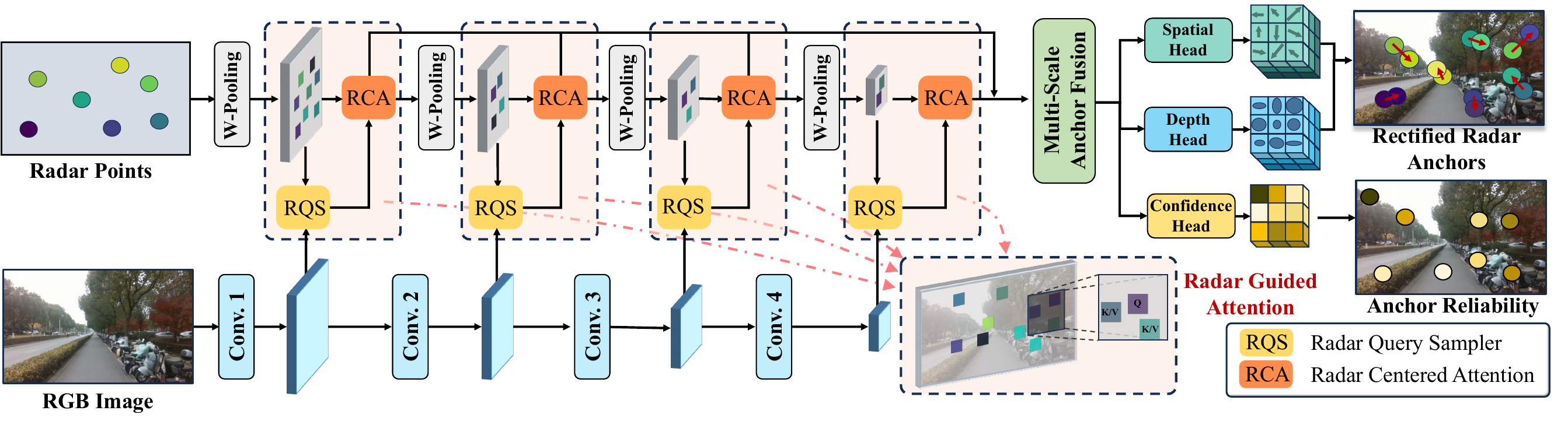}
    \caption{
    Image-conditioned radar anchor rectification.
    RQS and RCA aggregate local visual evidence around each
    radar candidate across multiple scales, while MAF predicts
    spatial and depth corrections with pointwise reliability
    to produce rectified temporal anchors.
    }
    \label{fig:anchor_rectification}
\end{figure*}

\subsection{Image-Conditioned Radar Anchor Rectification}
\label{sec:anchor_rectification}

Accumulated radar returns provide metric measurements but may contain
inaccurate projections and depths due to limited angular resolution,
multipath, clutter, and temporal inconsistency.
We therefore rectify each temporal anchor candidate using local image
evidence before propagation.

As shown in Fig.~\ref{fig:anchor_rectification}, the module comprises a
radar query sampler (RQS), radar-centered attention (RCA), multi-scale
anchor fusion (MAF), and three prediction heads.
RQS samples local image features, while RCA first models radar-neighborhood
consistency through self-attention and weighted pooling and then retrieves
compatible visual evidence through cross-attention.
MAF integrates the resulting multi-scale representations before three
prediction heads estimate the spatial offset, depth residual, and reliability.

For each candidate $a_i\in\mathcal A_t$, its depth and radar attributes
are encoded as
\begin{equation}
\mathbf f_i^r
=
\mathcal E_{\mathrm{rad}}([z_i,\mathbf r_i]).
\end{equation}
Meanwhile, the image encoder extracts a multi-scale feature pyramid:
\begin{equation}
\{\mathbf F^s\}_{s=1}^{S}
=
\mathcal E_{\mathrm{img}}(I_t),
\end{equation}
where $\mathbf p_i^s$ denotes $\mathbf p_i$ mapped to the coordinate
system of $\mathbf F^s$.

\paragraph{Radar Query Sampler.}
Directly sampling at $\mathbf p_i^s$ may provide unreliable visual
evidence when the radar projection is inaccurate.
At each scale, RQS therefore predicts $M$ radar-conditioned sampling
offsets:
\begin{equation}
\{\boldsymbol{\delta}_{i,m}^{s}\}_{m=1}^{M}
=
\mathcal H_{\mathrm{off}}^{s}(\mathbf f_i^r),
\end{equation}
and extracts the corresponding image features through bilinear
sampling:
\begin{equation}
\mathbf x_{i,m}^{s}
=
\operatorname{Bil}
\left(
\mathbf F^s,
\mathbf p_i^s+\boldsymbol{\delta}_{i,m}^{s}
\right),
\qquad m=1,\ldots,M.
\end{equation}
This local sampling allows visual evidence to be collected around the
initial projection without requiring explicit correspondence search.

\paragraph{Radar-Centered Attention.}
RCA first models the local consistency among radar candidates and then
retrieves visually compatible evidence through cross-modal attention.
For each candidate $a_i$, we collect a local radar neighborhood
$\mathcal N_i^r$ and form its radar tokens as
\begin{equation}
\mathbf G_i^r
=
\{\mathbf g_{i,j}^r\mid a_j\in\mathcal N_i^r\},
\qquad
\mathbf g_{i,j}^r
=
[\mathbf f_j^r,\phi(\mathbf p_j-\mathbf p_i)],
\end{equation}
where $\phi(\cdot)$ encodes the relative position with respect to the
center candidate.

Self-attention is first applied within the radar neighborhood:
\begin{equation}
\mathbf Z_i^r
=
\operatorname{Attn}
\left(
\mathbf G_i^r\mathbf W_q^r,
\mathbf G_i^r\mathbf W_k^r,
\mathbf G_i^r\mathbf W_v^r
\right).
\end{equation}
The resulting neighbor features are summarized by weighted pooling:
\begin{equation}
\bar{\mathbf f}_i^r
=
\sum_{a_j\in\mathcal N_i^r}
\beta_{i,j}\mathbf z_{i,j}^r,
\qquad
\boldsymbol{\beta}_i
=
\operatorname{softmax}
\left(
\mathcal H_{\mathrm{wp}}(\mathbf Z_i^r)
\right).
\end{equation}
This radar self-attention and weighted pooling characterize the local
measurement consistency while suppressing unreliable neighboring returns.

At each image scale, the pooled radar representation serves as the query,
while the image features sampled by RQS provide the keys and values:
\begin{equation}
\mathbf q_i^s
=
\mathbf W_q^s\bar{\mathbf f}_i^r,
\qquad
\mathbf k_{i,m}^s
=
\mathbf W_k^s\mathbf x_{i,m}^s,
\qquad
\mathbf v_{i,m}^s
=
\mathbf W_v^s\mathbf x_{i,m}^s.
\end{equation}
The cross-attention output is computed as
\begin{equation}
\alpha_{i,m}^{s}
=
\operatorname{softmax}_{m}\!\left(
\frac{(\mathbf q_i^s)^\top\mathbf k_{i,m}^s}{\sqrt d}
\right),
\quad
\bar{\mathbf f}_i^{\,s}
=
\sum_{m=1}^{M}\alpha_{i,m}^{s}\mathbf v_{i,m}^{s}.
\end{equation}
Finally, the radar and visual representations are fused to produce the
scale-specific anchor feature:
\begin{equation}
\tilde{\mathbf f}_i^s
=
\mathcal F_{\mathrm{rca}}^s
\left(
[\bar{\mathbf f}_i^r,\bar{\mathbf f}_i^{\,s}]
\right).
\end{equation}
\paragraph{Multi-Scale Anchor Fusion.}
Fine-scale features preserve local boundaries, whereas coarse-scale
features provide broader contextual information.
MAF integrates the representations from all scales while retaining the
original radar feature:
\begin{equation}
\mathbf h_i
=
\mathcal F_{\mathrm{maf}}
\left(
\tilde{\mathbf f}_i^1,\ldots,
\tilde{\mathbf f}_i^S,
\mathbf f_i^r
\right).
\end{equation}

Three lightweight heads predict the image-plane offset, depth residual,
and pointwise reliability:
\begin{equation}
\Delta\mathbf p_i=\mathcal H_{\mathrm{spa}}(\mathbf h_i),\;
\Delta z_i=\mathcal H_{\mathrm{dep}}(\mathbf h_i),\;
c_i=\sigma\!\left(\mathcal H_{\mathrm{conf}}(\mathbf h_i)\right).
\end{equation}
The rectified location and depth are obtained through residual updates:
\begin{equation}
\hat{\mathbf p}_i
=
\mathbf p_i+\Delta\mathbf p_i,
\qquad
\hat z_i
=
z_i+\Delta z_i.
\end{equation}
Finally, we retain $\mathbf f_i^a=\mathbf h_i$ and define the rectified
temporal anchor set as
\begin{equation}
\hat{\mathcal A}_t
=
\left\{
\hat a_i
=
(\hat{\mathbf p}_i,\hat z_i,c_i,\mathbf f_i^a)
\right\}_{i=1}^{N}.
\end{equation}
Rather than serving as a binary mask, $c_i$ continuously modulates each
anchor's contribution during subsequent dense propagation.

\subsection{Reliability-Aware Anchor Propagation}
\label{sec:anchor_propagation}

Although rectification improves the quality of temporal radar anchors,
they remain sparse and unevenly distributed.
Moreover, directly propagating all nearby anchors may spread residual
misalignment, clutter, and dynamic returns into dense image regions.
We therefore introduce reliability-aware anchor propagation, which
selects informative anchors for each target location and adaptively
aggregates their metric evidence before dense refinement.

\begin{figure}[t]
    \centering
    \includegraphics[width=\linewidth]{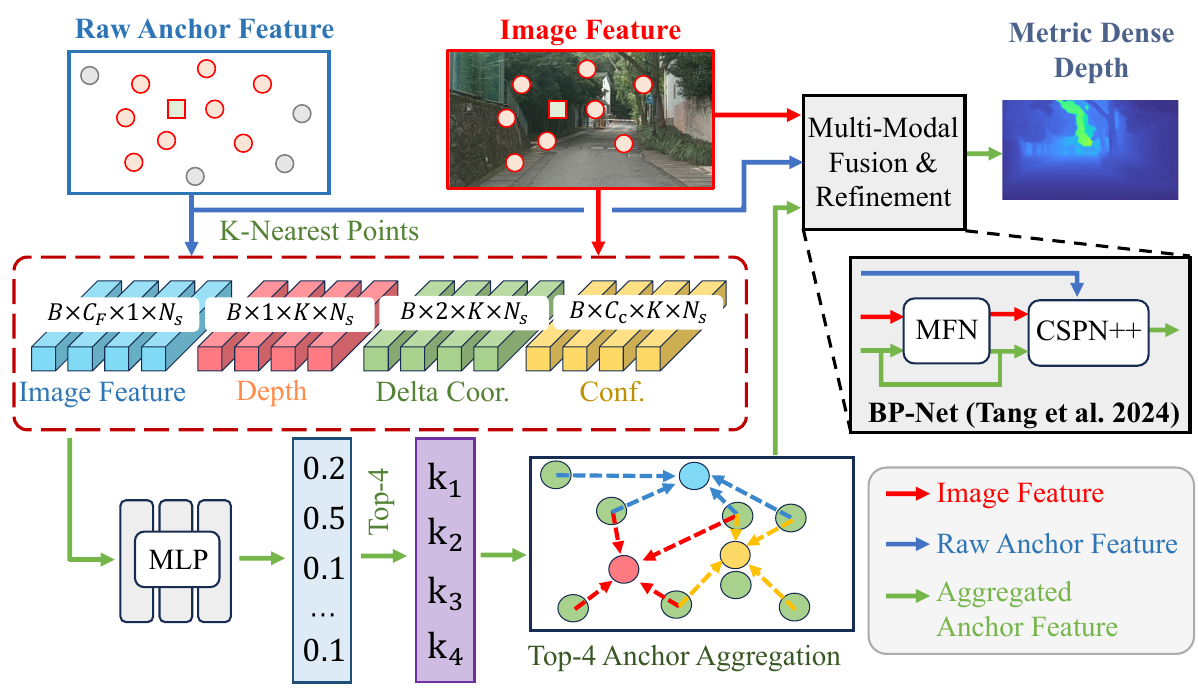}
    \caption{
    Reliability-aware anchor propagation.
    For each target location, spatially neighboring anchors are scored, and
    the top four are aggregated using reliability-modulated weights.
    }
    \label{fig:agg}
\end{figure}

\begin{table*}[t]
\centering
\caption{
Quantitative comparison on ZJU-4DRadarCam.
All plug-in methods use the same DPT-Hybrid \cite{ranftl2021vision}.
N/A denotes results not reported in the original paper.
Bold and underlined values indicate the category-wise best and
second-best results, respectively, while
\colorbox{bestpurple}{\strut best} and
\colorbox{secondyellow}{\strut second-best} indicate the corresponding
overall rankings.
}
\label{tab:zju_sota}
\setlength{\tabcolsep}{2.2pt}
\resizebox{\textwidth}{!}{%
\begin{tabular}{c l *{18}{c}}
\toprule
\multirow{2}{*}{Type} & \multirow{2}{*}{Method}
& \multicolumn{6}{c}{0--50m}
& \multicolumn{6}{c}{0--70m}
& \multicolumn{6}{c}{0--80m} \\
\cmidrule(lr){3-8} \cmidrule(lr){9-14} \cmidrule(lr){15-20}
&
& MAE$\downarrow$ & RMSE$\downarrow$ & iMAE$\downarrow$ & iRMSE$\downarrow$ & Rel$\downarrow$ & $\delta_1\uparrow$
& MAE$\downarrow$ & RMSE$\downarrow$ & iMAE$\downarrow$ & iRMSE$\downarrow$ & Rel$\downarrow$ & $\delta_1\uparrow$
& MAE$\downarrow$ & RMSE$\downarrow$ & iMAE$\downarrow$ & iRMSE$\downarrow$ & Rel$\downarrow$ & $\delta_1\uparrow$ \\
\midrule

\multirow{2}{*}{Plug-in}
& RadarCam ('24)
& \catsecond{1067.5} & \catsecond{2817.4} & \catsecond{10.5} & \catsecond{22.9} & \overallsecond{\catbest{0.087}} & \overallsecond{\catbest{0.920}}
& \catsecond{1157.0} & \catsecond{3117.7} & \catsecond{10.4} & \catsecond{22.9} & \catsecond{0.087} & \catsecond{0.921}
& \catsecond{1183.5} & \catsecond{3229.0} & \catsecond{10.4} & \catsecond{22.8} & \overallsecond{\catbest{0.090}} & \overallsecond{\catbest{0.920}} \\

& TacoDepth ('25)
& \overallsecond{\catbest{930.2}} & \overallsecond{\catbest{2477.3}} & \overallsecond{\catbest{9.3}} & \overallsecond{\catbest{20.8}} & \NA & \NA
& \overallbest{\catbest{983.1}} & \overallsecond{\catbest{2779.6}} & \overallsecond{\catbest{9.3}} & \overallsecond{\catbest{20.9}} & \overallsecond{\catbest{0.076}} & \overallsecond{\catbest{0.932}}
& \overallsecond{\catbest{1032.5}} & \overallsecond{\catbest{2850.3}} & \overallsecond{\catbest{9.4}} & \overallsecond{\catbest{20.9}} & \NA & \NA \\

\specialrule{0.08em}{1pt}{1pt}
\specialrule{0.08em}{1pt}{1pt}

\multirow{8}{*}{\shortstack{Indep-\\endent}}
& DORN ('21)
& 2210.2 & 4129.7 & 19.8 & 31.9 & 0.157 & 0.783
& 2402.2 & 4625.2 & 19.8 & 31.9 & 0.160 & 0.777
& 2447.6 & 4760.0 & 19.9 & 31.9 & 0.161 & 0.776 \\

& Singh ('23)
& 1785.4 & 3704.6 & 18.1 & 35.3 & 0.146 & 0.831
& 1932.7 & 4137.1 & 18.0 & 35.2 & 0.147 & 0.828
& 1979.5 & 4309.3 & 17.9 & 35.1 & 0.147 & 0.828 \\

& BP-Net ('24)
& 1404.8 & 2923.1 & 10.6 & \catsecond{21.2} & 0.104 & \catsecond{0.895}
& 1531.0 & 3259.6 & 10.5 & \catsecond{21.0} & 0.105 & \catsecond{0.892}
& 1568.2 & 3383.4 & \catsecond{10.4} & \catsecond{21.0} & 0.105 & \catsecond{0.892} \\

& XD-RC ('25)
& 1155.0 & 2863.0 & 11.7 & 24.6 & \catsecond{0.094} & 0.891
& 1248.0 & 3223.0 & 11.7 & 24.5 & \catsecond{0.095} & 0.890
& 1275.0 & 3341.0 & 11.7 & 24.5 & \catsecond{0.095} & 0.889 \\

& TacoDepth ('25)
& 1120.1 & 2686.7 & 12.8 & 25.0 & \NA & \NA
& 1181.8 & \catsecond{2906.3} & 12.7 & 24.9 & \NA & \NA
& 1201.1 & \catsecond{2990.7} & 12.7 & 24.9 & \NA & \NA \\

& JustDepth ('26)
& 1224.7 & 3157.5 & 12.8 & 25.2 & 0.096 & 0.888
& 1307.8 & 3479.0 & 12.7 & 25.2 & 0.096 & 0.888
& 1334.6 & 3618.1 & 12.7 & 25.2 & 0.096 & 0.888 \\

& SomeDepth ('26)
& \catsecond{1029.2} & \catsecond{2631.6} & \catsecond{10.4} & 22.8 & \NA & \NA
& \catsecond{1111.6} & 2946.9 & \catsecond{10.4} & 22.8 & \NA & \NA
& \catsecond{1137.2} & 3053.0 & \catsecond{10.4} & 22.8 & \NA & \NA \\

& \textbf{RbFT-Net}
& \overallbest{\catbest{917.4}} & \overallbest{\catbest{2473.4}} & \overallbest{\catbest{6.8}} & \overallbest{\catbest{16.9}} & \overallbest{\catbest{0.067}} & \overallbest{\catbest{0.944}}
& \overallsecond{\catbest{1001.0}} & \overallbest{\catbest{2740.8}} & \overallbest{\catbest{6.8}} & \overallbest{\catbest{16.8}} & \overallbest{\catbest{0.068}} & \overallbest{\catbest{0.943}}
& \overallbest{\catbest{1023.7}} & \overallbest{\catbest{2827.6}} & \overallbest{\catbest{6.7}} & \overallbest{\catbest{16.8}} & \overallbest{\catbest{0.068}} & \overallbest{\catbest{0.942}} \\

\bottomrule
\end{tabular}%
}
\end{table*}

\paragraph{Reliability-Guided Anchor Selection.}
Given the rectified temporal anchors $\hat{\mathcal A}_t$, we first retrieve the $K$ nearest
anchors for each target location $\mathbf p$ on the propagation feature map:
\begin{equation}
\mathcal N_K(\mathbf p)
=
\operatorname{KNN}
\left(
\mathbf p,\hat{\mathcal A}_t
\right).
\end{equation}

For each candidate $\hat a_i\in\mathcal N_K(\mathbf p)$, we construct
a target--anchor representation from the target image context, anchor
feature, rectified depth, and relative projected position:
\begin{equation}
\mathbf m_{p,i}
=
\left[
\mathbf F(\mathbf p),
\mathbf f_i^a,
\hat z_i,
\phi_p\!\left(\mathbf p-\hat{\mathbf p}_i\right)
\right],
\end{equation}
where $\mathbf F(\mathbf p)$ denotes the image feature at the target
location, and $\phi_p(\cdot)$ encodes the relative position between the
target and rectified anchor.

A lightweight MLP predicts a compatibility score:
\begin{equation}
s_i(\mathbf p)
=
\mathcal M_{\mathrm{sel}}
\left(
\mathbf m_{p,i}
\right).
\end{equation}
Rather than propagating all neighboring anchors, we retain the $K_s$
candidates with the highest compatibility scores:
\begin{equation}
\mathcal T_{K_s}(\mathbf p)
=
\underset{\hat a_i\in\mathcal N_K(\mathbf p)}
{\operatorname{TopK}}
\left(s_i(\mathbf p),K_s\right).
\end{equation}
We set $K_s=4$ in all experiments.

The selected anchors are adaptively weighted according to both their
target compatibility and predicted reliability:
\begin{equation}
w_i(\mathbf p)
=
\frac{
c_i\exp\!\left(s_i(\mathbf p)\right)
}{
\displaystyle
\sum_{\hat a_j\in\mathcal T_{K_s}(\mathbf p)}
c_j\exp\!\left(s_j(\mathbf p)\right)
}.
\end{equation}
The propagated anchor representation is then obtained as
\begin{equation}
\bar{\mathbf f}^{a}(\mathbf p)
=
\sum_{\hat a_i\in\mathcal T_{K_s}(\mathbf p)}
w_i(\mathbf p)
\mathcal E_a\!\left([\mathbf f_i^a,\hat z_i]\right),
\end{equation}
where $\mathcal E_a$ embeds the anchor feature together with its
rectified metric depth.

This two-stage strategy first restricts candidates by spatial proximity
and then selects visually compatible anchors through learned relevance.
The explicit reliability modulation suppresses residual erroneous
returns during propagation.

\paragraph{Multi-Modal Fusion and Refinement.}
After rectification and reliability-aware propagation, the temporal
anchors form the propagated feature map $\bar{\mathbf F}^{a}$.
We then adopt the lightweight MFN-CSPN++ architecture of
BP-Net~\cite{tang2024bilateral} for high-level multi-modal fusion
and dense refinement:
\begin{equation}
D_t
=
\mathcal R_{\mathrm{ref}}
\left(
\mathbf F,
\mathbf F_{\mathrm{rad}}^{t},
\bar{\mathbf F}^{a}
\right),
\end{equation}
where $\mathbf F_{\mathrm{rad}}^{t}$ preserves the directly observed
current-frame measurements, while $\bar{\mathbf F}^{a}$ provides
reliability-aware metric evidence from the rectified temporal anchors.

Unlike BP-Net, which directly aggregates the four nearest sparse
measurements, RbFT-Net performs reliability-aware aggregation over
rectified temporal anchors before lightweight MFN-CSPN refinement.

\subsection{Training Objectives}
\label{sec:training_objectives}

The network is trained end-to-end using LiDAR-projected depth supervision. The overall objective is
\begin{equation}
\mathcal L
=
\lambda_d\mathcal L_{\mathrm{depth}}
+\lambda_a\mathcal L_{\mathrm{anchor}}
+\lambda_c\mathcal L_{\mathrm{conf}}
+\lambda_p\mathcal L_{\mathrm{prop}}.
\end{equation}
Here, $\mathcal L_{\mathrm{depth}}$ supervises the final depth prediction,
while $\mathcal L_{\mathrm{anchor}}$ constrains the rectified anchor depths.
$\mathcal L_{\mathrm{conf}}$ provides error-aware supervision for anchor
reliability, and $\mathcal L_{\mathrm{prop}}$ encourages the propagation
module to favor depth-consistent anchors.
Detailed loss definitions and supervision construction are provided in
the supplementary material.

\begin{figure*}[t]
    \centering
    \includegraphics[width=\linewidth]{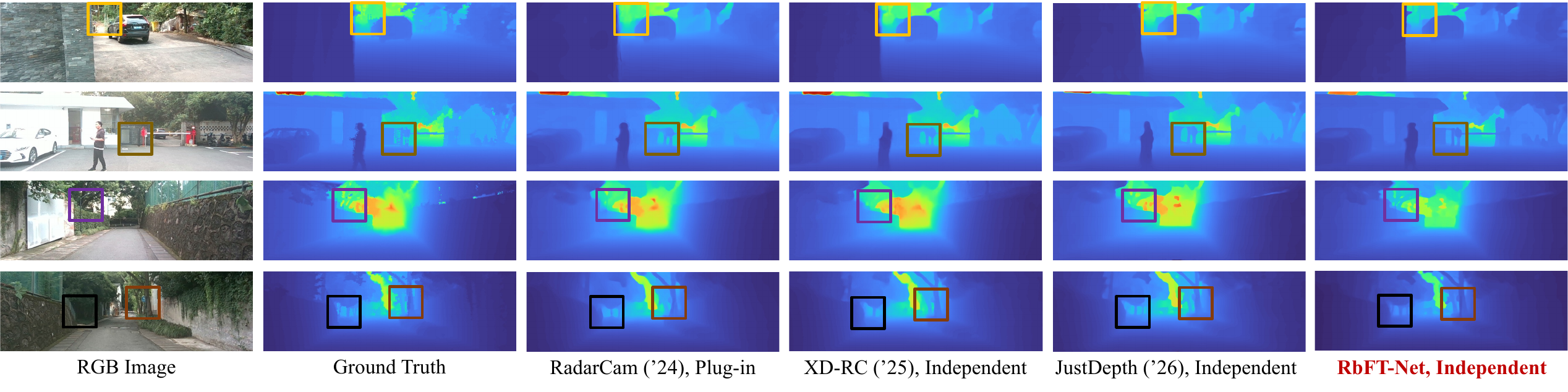}
    \caption{
    Qualitative comparison on ZJU-4DRadarCam.
    }
    \label{fig:zju_results}
\end{figure*}

\section{Experiments}
\subsection{Experimental Setup}
\label{sec:exp_setup}

\paragraph{Datasets.}
We evaluate RbFT-Net on ZJU-4DRadarCam~\cite{li2024radarcam} and a newly
collected dataset from a different 4D radar--camera platform.
Both provide synchronized radar, camera, and LiDAR data, with projected
radar as input and LiDAR depth as ground truth.

\paragraph{Compared Methods.}
We compare with representative plug-in
methods~\cite{li2024radarcam,wang2025tacodepth} and independent
methods~\cite{lo2021depth,singh2023depth,sun2025xd,
wang2025tacodepth,yun2026justdepth,hou2026selection}.
The evaluated plug-in methods use DPT-Hybrid~\cite{ranftl2021vision}
for auxiliary monocular depth.
We also adapt BP-Net~\cite{tang2024bilateral} to radar inputs as a
propagation baseline without anchor rectification or reliability modeling.

\paragraph{Evaluation Metrics.}
We report MAE and RMSE in millimeters, iMAE and iRMSE in
$1/\mathrm{km}$, and the unitless AbsRel and $\delta_1$ over valid
ground-truth pixels. Lower values are better except for $\delta_1$.

\paragraph{Implementation Details.}
RbFT-Net is trained at $288\times864$ on ZJU-4DRadarCam and
$288\times832$ on our dataset, using five accumulated radar frames by default.
For each target location, it retrieves $K=8$ rectified anchors and aggregates
the top $K_s=4$ based on learned compatibility scores.
Additional settings and dataset statistics are provided in the supplementary material.

\subsection{Comparison with State-of-the-Art Methods}
\paragraph{ZJU-4DRadarCam.}
Table~\ref{tab:zju_sota} reports the quantitative results on
ZJU-4DRadarCam.
RbFT-Net consistently achieves the best performance among
independent methods across all metrics and evaluation ranges.
Compared with the strongest independent baseline for each metric,
it reduces MAE by approximately 10\%, iMAE by about 35\%, and
iRMSE by over 25\%.
Without relying on the auxiliary DPT-Hybrid monocular depth model,
RbFT-Net remains competitive with, and in most reported settings surpasses, the plug-in TacoDepth configuration.
The qualitative comparisons in Fig.~\ref{fig:zju_results} show
cleaner depth predictions with sharper object boundaries and fewer
artifacts in distant regions.

\begin{table}[t]
\centering
\caption{
Evaluation on our newly collected dataset within the
0--70\,m range under different target-domain data regimes.
Zero-shot models are trained only on
ZJU-4DRadarCam; adaptation models are fine-tuned for three
epochs using 10\% of the target-domain training set.
}
\label{tab:collected_transfer}
\setlength{\tabcolsep}{1.6pt}
\renewcommand{\arraystretch}{1.05}
\resizebox{\columnwidth}{!}{%
\begin{tabular}{c l cccccc}
\toprule
Setting & Method
& MAE $\downarrow$
& RMSE $\downarrow$
& iMAE $\downarrow$
& iRMSE $\downarrow$
& Rel $\downarrow$
& $\delta_1$ $\uparrow$ \\
\midrule

\multirow{5}{*}{\shortstack{Standard\\100\%}}
& DORN
& 2243.2 & 4928.2 & 6.4 & 16.6 & 0.106 & 0.884 \\

& Singh
& 1824.6 & 4685.2 & 5.9 & 15.7
& 0.092 & 0.918 \\

& RadarCam
& 1644.9 & \catsecond{4120.7} & 5.7 & \catsecond{14.4}
& 0.078 & \catsecond{0.938} \\

& JustDepth
& \catsecond{1559.7} & 4229.3 & \catsecond{4.5} & 15.4
& \catsecond{0.072} & \catsecond{0.938} \\

& \textbf{RbFT-Net}
& \catbest{1430.4}
& \catbest{3622.1}
& \catbest{4.1}
& \catbest{12.4}
& \catbest{0.069}
& \catbest{0.941} \\

\midrule

\multirow{3}{*}{\shortstack{Zero-shot\\0\%}}
& RadarCam
& 9373.0 & \catsecond{13073.8} & 32.4 & 44.8
& 0.412 & 0.284 \\

& JustDepth
& \catsecond{9286.3} & 13431.7 & \catsecond{27.9} & \catsecond{38.2}
& \catsecond{0.360} & \catsecond{0.331} \\

& \textbf{RbFT-Net}
& \catbest{7185.9} & \catbest{10692.7} & \catbest{21.5} & \catbest{31.1}
& \catbest{0.319} & \catbest{0.421} \\

\midrule

\multirow{3}{*}{\shortstack{Adaptation\\10\%}}
& RadarCam
& \catsecond{3416.7} & \catsecond{5792.0} & \catsecond{8.0} & \catsecond{16.1} & \catsecond{0.137} & \catsecond{0.835} \\

& JustDepth
& 4082.6 & 6821.5& 11.2 & 27.1 & 0.186 & 0.752 \\

& \textbf{RbFT-Net}
& \catbest{2312.3} & \catbest{4696.8} & \catbest{6.4} & \catbest{15.0} & \catbest{0.105} & \catbest{0.895} \\

\bottomrule
\end{tabular}%
}
\end{table}

\begin{table}[t]
\centering
\caption{
0--70\,m accuracy--efficiency comparison on ZJU-4DRadarCam.
FPS is measured on an NVIDIA RTX PRO 5000.
RbFT-Net uses five radar frames; all other methods use one.
Plug-in parameter counts include auxiliary depth models.
}
\label{tab:efficiency}
\setlength{\tabcolsep}{3.0pt}
\renewcommand{\arraystretch}{1.08}
\small
\resizebox{\columnwidth}{!}{%
\begin{tabular}{llccccc}
\toprule
Type & Method
& Params (M)$\downarrow$
& FPS$\uparrow$
& MAE$\downarrow$
& RMSE$\downarrow$
& $\delta_1\uparrow$ \\
\midrule

\multirow{2}{*}{Plug-in}
& RadarCam \cite{li2024radarcam}
& 156.26 & 12.04
& 1157.0 & 3117.7 & 0.921 \\

& TacoDepth \cite{wang2025tacodepth}
& 137.25 & \NA
& \best{983.1}
& \second{2779.6}
& \second{0.932} \\

\specialrule{0.08em}{1pt}{1pt}
\specialrule{0.08em}{1pt}{1pt}

\multirow{4}{*}{\shortstack{Indep-\\endent}}
& DORN \cite{lo2021depth}
& 107.88 & 24.74
& 2402.2 & 4625.2 & 0.777 \\

& BP-Net \cite{tang2024bilateral}
& 89.87 & 12.04
& 1531.0 & 3259.6 & 0.892 \\

& JustDepth \cite{yun2026justdepth}
& \best{16.07}
& \best{110.06}
& 1307.8 & 3479.0 & 0.888 \\

& \textbf{RbFT-Net (Ours)}
& \second{44.20}
& \second{44.88}
& \second{1001.0}
& \best{2740.8}
& \best{0.943} \\

\bottomrule
\end{tabular}%
}
\end{table}

\paragraph{Evaluation on the Newly Collected Dataset.}
Table~\ref{tab:collected_transfer} compares the methods under standard
training, zero-shot transfer, and limited-data adaptation.
With the full target-domain training set, RbFT-Net achieves the strongest
overall performance.
For cross-platform evaluation, ZJU-4DRadarCam serves as the source domain
and the newly collected dataset as the target domain.
Under zero-shot transfer, RbFT-Net consistently outperforms the compared
methods despite differences in sensing platforms and scene distributions,
reducing MAE by 22.6\% relative to JustDepth and RMSE by 18.2\% relative
to RadarCam.
When fine-tuned for three epochs using the same 10\% target-domain subset,
RbFT-Net again achieves the best results across all metrics, demonstrating
effective adaptation under limited target-domain supervision.

\paragraph{Accuracy and Efficiency.}
Table~\ref{tab:efficiency} compares representative methods under their
default input configurations. RbFT-Net uses only about \textbf{1/3}
of the parameters of complete plug-in pipelines, whose parameter counts
include the auxiliary 123M-parameter DPT-Hybrid, while achieving
the best overall accuracy. Despite processing five radar frames, it runs
in real time at 44.88 FPS. Table~\ref{tab:frame_ablation} further
shows that directly extending competing methods to multiple frames yields
only limited accuracy gains, demonstrating the favorable
accuracy--efficiency trade-off of RbFT-Net.

\subsection{Ablation Studies}
\label{sec:ablation}

We conduct ablation studies on ZJU-4DRadarCam under the
0--70\,m evaluation range to analyze the contributions of the
proposed components and the effect of temporal radar input.

\paragraph{Component Analysis.}
Table~\ref{tab:component_ablation} evaluates the proposed rectification
and propagation designs.
With reliability estimation retained, spatial and depth rectification
individually reduce RMSE by 258.2 and 180.2~mm, respectively, and their
combination provides further improvement.
Given fully rectified anchors, learned propagation outperforms direct
4-NN aggregation, while confidence guidance further reduces RMSE from
2980.1 to 2740.8~mm.
These results demonstrate the complementary effects of anchor
rectification and reliability-aware propagation.

\begin{table}[t]
\centering
\caption{
Component ablation on ZJU-4DRadarCam.
All variants use five radar frames.
\emph{4-NN} directly aggregates the four nearest anchors, whereas
\emph{Learned} denotes feature-based propagation.
}
\label{tab:component_ablation}
\setlength{\tabcolsep}{2.3pt}
\renewcommand{\arraystretch}{1.06}
\small
\resizebox{\columnwidth}{!}{%
\begin{tabular}{ccc| c| cccc}
\toprule
\multicolumn{3}{c|}{Rectification}
& Anchor
& \multicolumn{4}{c}{Metrics [0--70\,m]} \\
\cmidrule(lr){1-3}
\cmidrule(lr){5-8}
Spatial
& Depth
& Conf.
& Propagation
& MAE $\downarrow$
& RMSE $\downarrow$
& Rel $\downarrow$
& $\delta_1 \uparrow$ \\
\midrule

           &            & \checkmark & Learned
& 1287.1 & 3064.6 & 0.087 & 0.920 \\

\checkmark &            & \checkmark & Learned
& 1055.5 & 2806.4 & 0.071 & 0.938 \\

           & \checkmark & \checkmark & Learned
& 1135.7 & 2884.4 & 0.077 & 0.929 \\

\midrule

\checkmark & \checkmark &            & 4-NN
& 1355.7 & 3062.6 & 0.096 & 0.904 \\

\checkmark & \checkmark &            & Learned
& 1207.3 & 2980.1 & 0.081 & 0.923 \\

\checkmark & \checkmark & \checkmark & Learned
& \textbf{1001.0}
& \textbf{2740.8}
& \textbf{0.068}
& \textbf{0.943} \\

\bottomrule
\end{tabular}%
}
\end{table}

\begin{table}[t]
\centering
\caption{Effect of the number of radar frames within the
0--70\,m range in RMSE (mm). Radar points denote the average
number of valid projected returns per sample.}
\label{tab:frame_ablation}
\resizebox{\columnwidth}{!}{%
\setlength{\tabcolsep}{5.0pt}
\renewcommand{\arraystretch}{1.08}
\small
\begin{tabular}{lccccc}
\toprule
\multirow{2}{*}{Method}
& \multicolumn{5}{c}{Number of Radar Frames} \\
\cmidrule(lr){2-6}
& 1 & 2 & 3 & 5 & 7 \\
\midrule

Avg. radar points
& 493.5 & 987.2 & 1480.9 & 2468.0 & 3455.3 \\
\midrule

RadarCam \cite{li2024radarcam}
& 3117.7 & 3104.1 & 3092.4 & \textbf{3071.9} & \underline{3085.1} \\

BP-Net \cite{tang2024bilateral}
& 3259.6 & 3172.4 & \underline{3135.8} & \textbf{3121.3} & 3156.8 \\



\textbf{RbFT-Net (Ours)}
& 2987.3 & 2841.5 & 2794.0
& \best{2740.8} & \second{2760.2} \\

\bottomrule
\end{tabular}%
}
\end{table}

\begin{figure}[t]
    \centering
    \includegraphics[width=\linewidth]{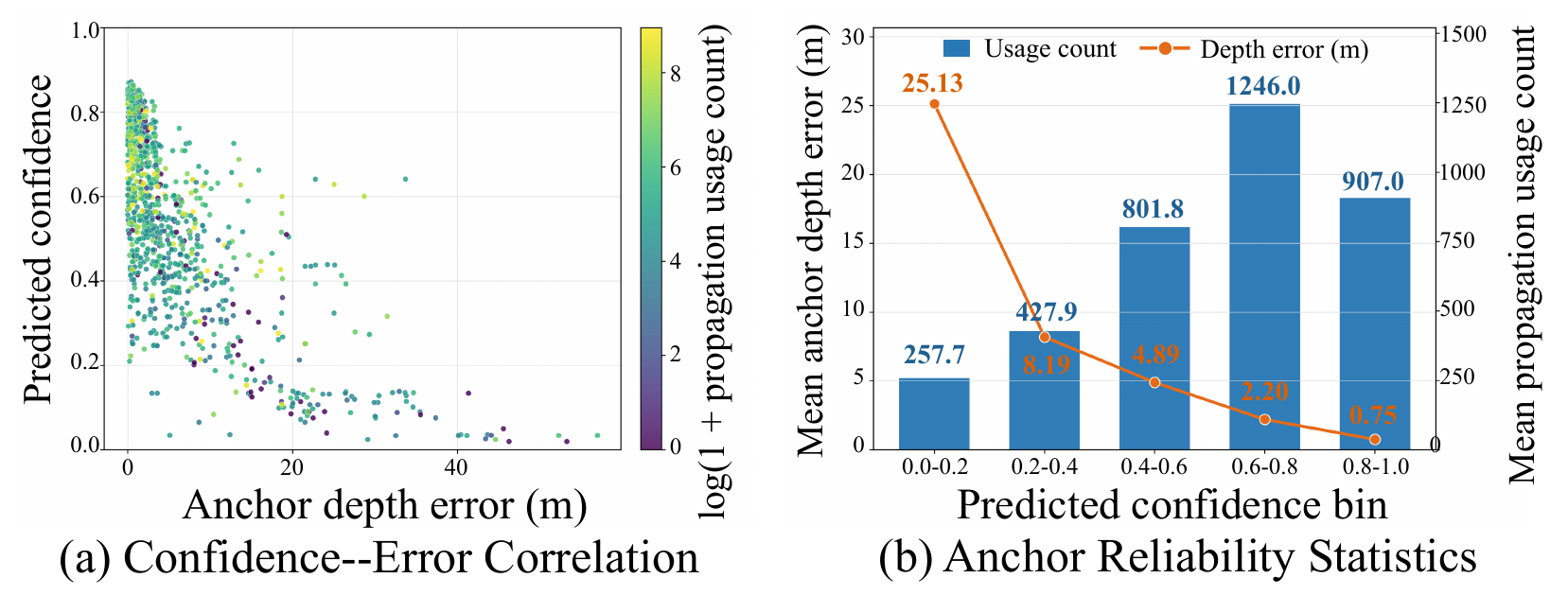}
    \caption{
    Reliability analysis of rectified radar anchors.
    Higher reliability corresponds to lower depth error and greater
    propagation usage.
    }
    \label{fig:confidence_analysis}
\end{figure}

\begin{figure}[t]
    \centering
    \includegraphics[width=\linewidth]
    {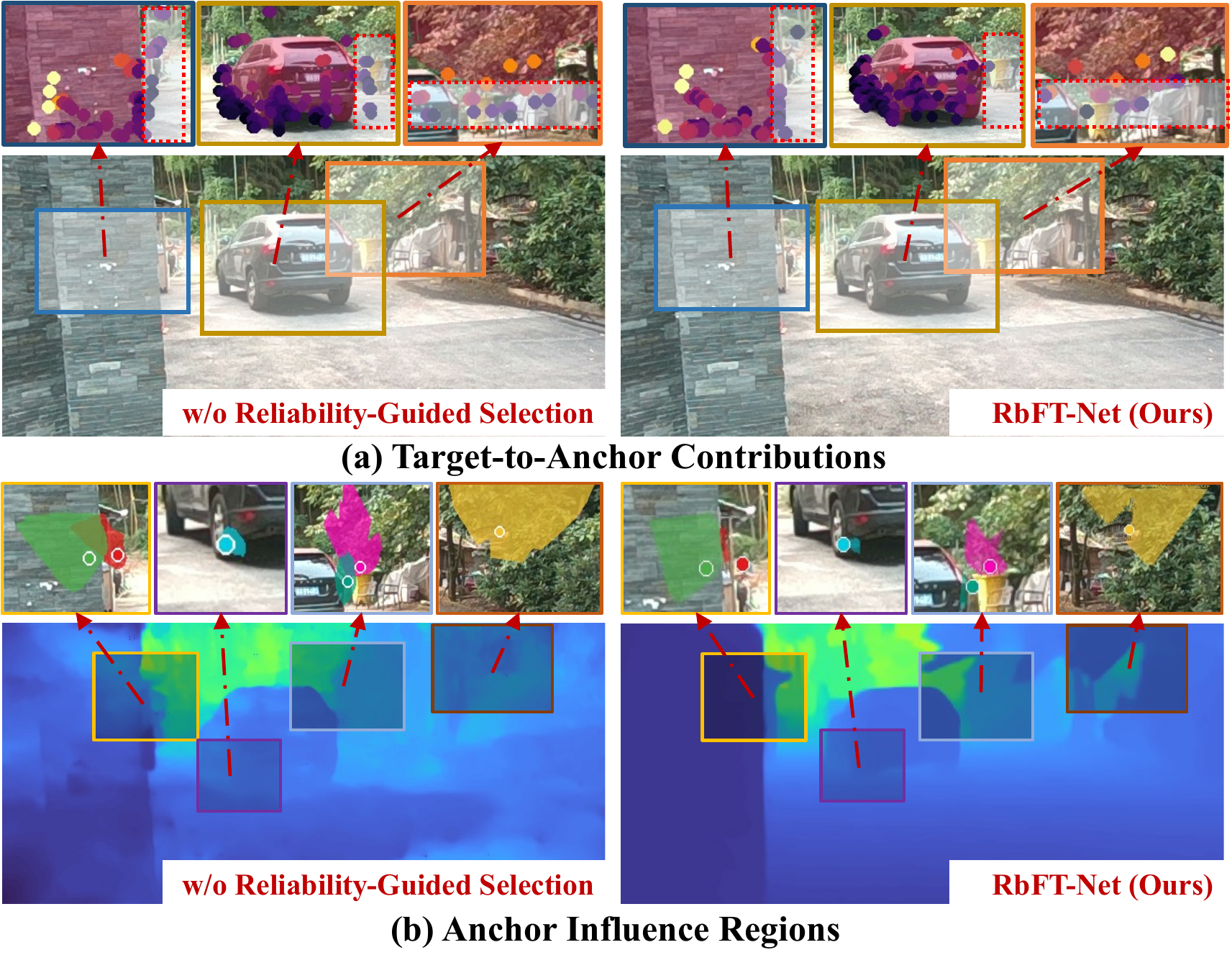}
    \caption{
    Visualization of learned target--anchor associations.
    (a) Anchors contributing to selected target regions.
    (b) Target regions influenced by selected anchors.
    Reliability-guided selection reduces propagation across unrelated
    structures and object boundaries.
    }
    \label{fig:anchor_consistency}
\end{figure}

\paragraph{Effect of Temporal Input.}
Table~\ref{tab:frame_ablation} compares methods retrained for each temporal
window under the same direct-accumulation setting.
Despite the near-linear increase in projected points, RadarCam and BP-Net
gain only modestly from additional frames, whereas RbFT-Net reduces RMSE
by 246.5~mm from one to five frames, demonstrating the benefit of rectifying
temporal returns before propagation.
The slight degradation at seven frames suggests increased misalignment and
noise, while processing more radar points also incurs additional computational
cost.
We therefore use five frames by default as a practical trade-off among
observation density, temporal noise, and efficiency.

\subsection{Analysis and Visualization}
\label{sec:analysis_vis}

\paragraph{Reliability Analysis.}
Fig.~\ref{fig:confidence_analysis} examines the relationship among
predicted reliability, anchor depth error, and propagation usage.
Both the point-wise correlation and binned statistics show that anchor
depth error generally decreases with increasing reliability.
High-reliability anchors also tend to contribute more frequently to the
learned target--anchor associations.
This relationship is not absolute, since anchor selection additionally
depends on target--anchor compatibility and the relative quality of nearby
candidates.
Overall, these results indicate that the predicted reliability provides
an effective cue for identifying accurate and informative rectified
anchors during dense depth propagation.

\paragraph{Structure-Consistent Propagation.}
Fig.~\ref{fig:anchor_consistency} visualizes target--anchor associations
from two complementary perspectives.
Fig.~\ref{fig:anchor_consistency}(a) shows the anchors aggregated for
selected target locations, while Fig.~\ref{fig:anchor_consistency}(b)
shows the target regions influenced by selected anchors.
Without reliability-guided selection, the four nearest anchors are
directly aggregated based on spatial proximity, often propagating
information across unrelated structures and object boundaries.
In contrast, RbFT-Net selects reliable anchors compatible with each
target location, favoring depth evidence from the same local surface
and reducing propagation across object boundaries.
This structure-consistent behavior emerges without explicit semantic
supervision.

\section{Conclusion}

This paper presented RbFT-Net, a compact independent framework for
multi-frame 4D radar--camera depth completion.
Its rectify-before-fuse design corrects noisy temporal anchors before
reliability-aware propagation and multi-modal fusion.
By modeling anchor locations, depths, and reliability, RbFT-Net exploits
denser temporal cues while suppressing unreliable propagation.
Experiments on a public benchmark and a newly collected dataset, including
cross-platform transfer and limited-data adaptation, demonstrate its
robustness across radar--camera platforms.

\bigskip

\bibliography{aaai2027}


\end{document}